\documentclass[10pt,letterpaper]{article}

\usepackage{cogsci}
\usepackage{graphicx}
\usepackage{booktabs}
\usepackage{amsmath}     
\usepackage{amssymb}     
\usepackage{bm}          
\usepackage{mathtools}   
\usepackage{xcolor}

\cogscifinalcopy 

\usepackage{pslatex}
\usepackage{apacite}
\usepackage{float} 

\title{Teasing Apart Architecture and Initial Weights \\ as Sources of Inductive Bias in Neural Networks}
 
\author{
    {\large \bf Gianluca Bencomo}$^1$,
    {\large \bf Max Gupta}$^1$,
    {\large \bf Ioana Marinescu}$^{2}$,
    {\large \bf R. Thomas McCoy}$^3$,
    {\large \bf Thomas L. Griffiths}$^{1,4}$ \\ 
    \texttt{gb5435@princeton.edu, mg7411@princeton.edu, im2178@nyu.edu, tom.mccoy@yale.edu, tomg@princeton.edu} \\
    $^1$Department of Computer Science, Princeton University \\
    $^2$Center for Data Science, New York University \\
    $^3$Department of Linguistics and Wu Tsai Institute, Yale University \\
    $^4$Department of Psychology, Princeton University
}

\begin{document}

\maketitle

\begin{abstract}
Artificial neural networks can acquire many aspects of human knowledge from data, making them promising as models of human learning.
But what those networks can learn depends upon their inductive biases -- the factors other than the data that influence the solutions they discover -- and the inductive biases of neural networks remain poorly understood, limiting our ability to draw conclusions about human learning from the performance of these systems. Cognitive scientists and machine learning researchers often focus on the architecture of a neural network as a source of inductive bias. In this paper we explore the impact of another source of inductive bias -- the initial weights of the network -- using meta-learning as a tool for finding initial weights that are adapted for specific problems. We evaluate four widely-used architectures -- MLPs, CNNs, LSTMs, and Transformers -- by meta-training 430 different models across three tasks requiring different biases and forms of generalization. We find that meta-learning can substantially reduce or entirely eliminate performance differences across architectures and data representations, suggesting that these factors may be less important as sources of inductive bias than is typically assumed. When differences are present, architectures and data representations that perform well without meta-learning tend to meta-train more effectively. Moreover, all architectures generalize poorly on problems that are far from their meta-training experience, underscoring the need for stronger inductive biases for robust generalization.

\textbf{Keywords:} inductive biases; neural networks; meta-learning
\end{abstract}

\section{Introduction}

Artificial neural networks have been used to explain how aspects of human knowledge that have been claimed to depend upon an extensive degree of innateness -- such as elements of language -- might be learned from data by systems that do not have strong built-in assumptions \cite<e.g.,>[]{rumelhartm86}. These networks offer a new perspective on central questions in cognitive science, such as what information we need to assume is innate to human learners \cite{elman1996rethinking}. In machine learning, a parallel set of questions focuses on the {\em inductive biases} of neural networks -- defined as those factors other than the data that influence the solutions that they find \cite{mitchell97}. The convergence of these literatures offers an opportunity to explore different ways in which innate knowledge might be implicitly expressed in artificial neural networks. 

Different neural network architectures display different inductive biases. For instance, one clear signature of inductive bias is the amount of data needed to learn a task, and convolutional neural networks can learn image classification tasks from less data than multi-layer perceptrons \cite{chen2021review}. In addition to network architecture, however, recent work has highlighted the importance of a network’s initial weights as a source of inductive bias \cite{finn2017model}. Specifically, techniques based on meta-learning can optimize the initial weights of a neural network (leaving the architecture unchanged) in ways that enable the network to learn new tasks from far less data than it would require using standard, randomly-selected initial weights. For instance, a network with meta-learned initial weights can learn new linguistic rules from just 100 examples, compared to the roughly 20,000 examples needed by the same architecture with non-meta-learned initial weights \cite{mccoy2020universal}. Such meta-learning results show that a given neural network architecture can realize very different inductive biases thanks to the flexibility afforded by the initial weights.

Here we consider this flexibility from the opposite direction: can a given inductive bias be realized equally well in very different network architectures? This question directly engages with the issue of whether architecture or initial weights provide a better focus for understanding the innate constraints on learning implicitly instantiated in a neural network. Prior work using meta-learning typically makes comparisons within a fixed architecture, comparing a version of that architecture with meta-learned initial weights to a version with randomly-selected initial weights. These comparisons make it clear that the initial weights afford a substantial degree of flexibility, but they leave open the question of whether that flexibility is extensive enough to override the influence of architecture such that a given inductive bias could be realized equally well in different architectures.

To address this, we explore several inductive biases, investigating how compatible each inductive bias is with different types of network architectures and data representations. We consider four widely-used, general-purpose neural architectures—multilayer perceptrons (MLPs; \citeNP{rosenblatt1962principles}), convolutional neural networks (CNNs; \citeNP{lecun1998gradient}), long short-term memory networks (LSTMs; \citeNP{hochreiter1997long}), and Transformers \cite{vaswani2017attention}—with variations in depth and width, meta-training a total of 430 models. To establish baselines where differences across architectures and data representations should be more pronounced—free from task-specific biases introduced by meta-learning—we compare these meta-trained models to the same architectures trained under typical regimes, starting from random initialization and optimizing along that trajectory. This design enables us to isolate how much of the performance variation can be attributed to architectural and data representation choices, as opposed to the learning processes that are agnostic to those choices.

Across both data representation and architecture, we observe substantial performance differences when models are trained using the usual approach of setting the initial weights randomly. However, introducing meta-learned inductive biases reduces, and in some cases completely eliminates, these differences, demonstrating that a given inductive bias can be instantiated in multiple, disparate architectures. Interestingly, architectures and data representations that perform well under random initialization also tend to meta-train more effectively, suggesting that some residual biases remain important for certain tasks. In few-shot learning, for example, models that excel without meta-learning are less sensitive to shifts in the training task distribution. Despite this, when models are required to learn tasks that lie far outside the distribution of tasks they encountered during meta-training, all architectures—regardless of inductive bias—fail catastrophically. This highlights that these general-purpose architectures may require stronger inductive biases for more robust forms of generalization but remain general enough to realize a wide range of biases given appropriate choices for the initial weights and learning rate.

\section{Background}
Inductive biases—the assumptions that guide learning—can manifest through the choice of model architecture, data representation, error metric, and training algorithm \cite{baxter2000model}. In this work, we investigate the extent to which model architecture and data representation influence performance outcomes after optimizing the initial weights and learning rate through meta-learning. This section introduces the kinds of biases inherent to the neural architectures we explore and addresses how meta-learning distills task-specific knowledge of the learning problem into the training algorithm.

\subsection{Inductive Biases across Neural Architectures}
\subsubsection{Multi-Layer Perceptrons (MLPs)}
MLPs \cite{rosenblatt1962principles} can approximate any function given sufficient depth and width \cite{hornik1989universal} but make no explicit assumption about the structure of the input data beyond static size. The lack of built-in equivariances make them highly sensitive to nearly all spatial and temporal variations. All-to-all connections between layers imply global feature mixing and deeper layers can capture progressively more abstract representations.

\subsubsection{Convolutional Neural Networks (CNNs)}
CNNs \cite{lecun1998gradient} were designed with an explicit bias towards grid-structured data such as images. Convolutions with shared weights prioritize spatially local relationships and ensure translation equivariance. Pooling layers provide partial robustness to variations in scale, though CNNs generally lack inherent rotation or scale equivariances. CNNs preserve spatial order, making them sensitive to input permutations. Like MLPs, they build hierarchical features, with deeper layers capturing abstract patterns composed of simpler ones. These representations often resemble those in the mammalian visual cortex \cite{yamins2016sensory}.

\subsubsection{Long Short-Term Memory Networks (LSTMs)}
LSTMs \cite{hochreiter1997long} were designed to capture both long- and short-term dependencies in sequences by maintaining an internal state that tracks temporal dynamics. Input, forget, and output gates regulate the flow of information, enabling the model to selectively retain or discard data. LSTMs rely on a memory structure with a bottleneck defined by the size of the hidden and cell states. They assume order matters, making them sensitive to input permutations. Sequential processing encodes position-awareness. Stacked layers allow LSTMs to capture hierarchical temporal patterns.

\subsubsection{Transformers (TFs)}  
Transformers \cite{vaswani2017attention} are designed to capture both local and global dependencies in sequences using a self-attention mechanism. Unlike LSTMs, which process input sequentially, Transformers compute attention over all input positions simultaneously. Self-attention enables the model to dynamically focus on relevant parts of the input, giving it direct access to long-range dependencies instead of through memory as in the LSTM. Transformers require explicit positional encodings since they lack an inherent sense of order. Stacked attention and feedforward layers enable the learning of hierarchical patterns, similar to deep CNNs when the input image is patched as is done with Vision Transformers \cite{dosovitskiy2021vit}.

\subsection{Meta-Learning}
We adopt a meta-learning approach called Meta-SGD \cite{li2017meta}, which learns a model initialization, learning rate, and update direction that solves a set of tasks in a fixed number of steps. Meta-SGD is an extension of Model-Agnostic Meta-Learning (MAML; \citeNP{finn2017model}), which only learns the model initialization. This enhanced flexibility allows for a more complete training algorithm that can quickly adapt to new, unseen tasks with minimal data. During meta-training, the goal is to learn a model initialization, learning rate, and update direction that extract shared structure across tasks and embed inductive biases into the model outside what is explicitly defined by architecture or data representation.

Formally, Meta-SGD aims to find parameters $\theta$ and $\alpha$ that minimize the expected test loss across tasks:
\begin{align*}
    \min_{\theta, \alpha} \, \mathbb{E}_{\tau \sim p(\tau)} \left[ \mathcal{L}_{\text{Test}} \left( \theta' \right) \right], \quad 
    \text{where} \quad \theta' = \theta - \alpha \odot \nabla_{\theta} \mathcal{L}_{\text{Train}}(\theta).
\end{align*}
Here, $p(\tau)$ denotes the task distribution, $\mathcal{L}_{\text{Train}}$ is the training loss for an individual task given a support set, and $\mathcal{L}_{\text{Test}}$ is the test loss given a query set. The parameter $\theta$ represents the model's initial weights, while $\alpha$ represents the learned task-specific learning rate and step direction for adaptation.

Prior work has demonstrated that meta-learning can embed inductive biases beyond those defined by a model's architecture. For example, \citeA{snell2023metalearned} showed how meta-learned neural circuits can perform complex, task-specific probabilistic reasoning by distilling the biases required to perform nonparametric Bayesian inference. Similarly, meta-learning has been applied to address the problem of catastrophic forgetting, a major challenge in online learning, by helping neural networks retain knowledge across tasks \cite{javed2019meta}. For tasks that require producing novel combinations from known components, \citeA{lake2023human} demonstrated meta-learning's ability to distill human-like compositional skills into neural networks, despite \citeA{fodor1988connectionism} famously arguing that artificial neural networks lacked this capacity. While these approaches focus on learning capabilities that neural networks may not inherently possess, our work investigates the extent to which specific inductive biases can be encoded and expressed within different architectures through meta-learning; see \citeA{abnar2020transferring} for a different approach that encourages similarity between architectures by directly training one architecture to reproduce the outputs from another, rather than having different architectures meta-learn from the same task distribution.

\section{Approach}
We evaluate two types of scenarios to address constraints between neural architectures, data representation, and training algorithms. First, we assess the best-case scenario for meta-learning, where test tasks are fully in-distribution and ample meta-training data is provided. Second, we test more challenging conditions, with test and training tasks from different distributions and limited meta-training data, where other sources of bias are more persistent. For both cases, we also compare performance against randomly-initialized baselines to highlight the impact of architectural bias without meta-learned adaptations. We follow a consistent meta-learning procedure across three tasks, defining variations in depth and width for each architecture, selecting models based on performance on a meta-validation set, meta-training each architecture using Meta-SGD, and testing against controlled baselines.

\begin{table}[t]
    \centering
    \caption{Hyperparameter search space for each architecture.}
    \label{tab:hyperparams}
    \begin{tabular}{lcc}
        \toprule
        \textbf{Architecture} & \textbf{Number of Layers} & \textbf{Hidden Width} \\
        \midrule
        MLP         & $\{2, 4, 6, 8\}$  & $\{8, 16, 32, 64\}$ \\
        CNN         & $\{2, 4, 6, 8\}$  & See Table~\ref{tab:hyperparams_cnn}. \\
        LSTM        & $\{1, 2, 3, 4\}$  & $\{8, 16, 32, 64\}$ \\
        Transformer & $\{1, 2, 3, 4\}$  & $\{8, 16, 32, 64\}$ \\
        \bottomrule
    \end{tabular}
\end{table}

\begin{table}[t]
    \centering
    \caption{Hidden Widths for CNN.}
    \label{tab:hyperparams_cnn}
    \begin{tabular}{lc}
        \toprule
        \textbf{\# Layers} & \textbf{Hidden Widths} \\
        \midrule
        $2$ & $\{(2^n, 2^{n-1}) \,|\, n \in \{4, 5, 6, 7\}\}$ \\
        $4$ & $\{(2^n, 2^{n-1}, 2^{n-2}, 2^{n-3}) \,|\, n \in \{4, 5, 6, 7\}\}$ \\
        $6$ & Same as for $4$ layers \\
        $8$ & Same as for $4$ layers \\
        \bottomrule
    \end{tabular}
\end{table}

\subsection{Neural Architectures}

To isolate the core inductive biases, we remove non-essential components, such as dropout. MLPs maintain a fixed hidden width across all layers, with batch normalization and ReLU activation applied after each hidden layer. CNNs use $3 \times 3$ kernels with a stride of 1 and zero-padding. Each convolutional layer is followed by batch normalization, a ReLU activation, and average pooling with a stride of 2. CNNs begin with either 2 or 4 convolutional layers (depending on the architecture depth) followed by fully connected layers with a fixed hidden width equal to the dimensionality of the final, flattened output of the convolution layers. LSTMs follow the original implementation in \citeA{hochreiter1997long} but use pre-layer normalization and a projection layer to align input and hidden dimensions. Transformers \cite{vaswani2017attention} use sinusoidal positional encoding, four attention heads, and a feedforward network with a dimensionality twice the hidden size. Pre-layer normalization and a projection layer are both used. There are 16 variations of each architecture, with ranges over depth, width, and parameter counts that are comparable (see Table~\ref{tab:hyperparams}).

\subsection{Meta-Training and Sampling Architectures}

We meta-train each architecture using Meta-SGD \cite{li2017meta}, with AdamW \cite{loshchilovdecoupled} as the outer optimizer. We set the learning rate to 0.001 and weight decay to 0.01. For all 16 variations of each architecture, we perform 10,000 episodes of meta-training. After this initial phase, we select the best-performing architecture based on its performance on a meta-validation set filled with 100 unseen training tasks and then continue optimizing the selected architecture to convergence. We repeat for 10 random seeds. This produces 10 independent samples of meta-learned weights and architectures for each architecture class. We meta-train with batches of 4 tasks and use 1 adaptation step throughout.

\begin{table*}[t]
    \centering
    \caption{Average accuracy for the concept learning task across different architectures, data types, and support set sizes. The table compares performance under meta-learning (\textbf{M1}) and random initialization conditions with 1, 10, and 200 steps of AdamW (\textbf{R1}, \textbf{R10}, \textbf{R200}). All 95\% confidence intervals (CIs) are below 0.01.}
    \label{tab:concept}
    \begin{tabular}{llcccc|cccc|cccc}
        \toprule
        & & \multicolumn{4}{c}{$n_{\text{support}} = 5$} & \multicolumn{4}{c}{$n_{\text{support}} = 10$} & \multicolumn{4}{c}{$n_{\text{support}} = 15$} \\
        \cmidrule(lr){3-6} \cmidrule(lr){7-10} \cmidrule(lr){11-14}
        \textbf{Arch.} & \textbf{Data} & \textbf{R1} & \textbf{R10} & \textbf{R200} & \textbf{M1} & \textbf{R1} & \textbf{R10} & \textbf{R200} & \textbf{M1} & \textbf{R1} & \textbf{R10} & \textbf{R200} & \textbf{M1} \\
        \midrule
        MLP & Image        & 0.499 & 0.551 & 0.635 & 0.823 & 0.515 & 0.613 & 0.775 & 0.935 & 0.489 & 0.616 & 0.859 & 0.955 \\
        CNN & Image        & 0.498 & 0.588 & 0.717 & 0.842 & 0.518 & 0.633 & 0.842  & 0.937 & 0.484 & 0.632 & 0.887 & 0.961 \\
        LSTM & Image      & 0.521 & 0.623 & 0.715 & 0.849 & 0.519 & 0.661 & 0.818 & 0.946 & 0.501 & 0.667 & 0.871 &  0.964 \\
        TF & Image        & 0.521 & 0.638 & 0.716 & 0.858 & 0.515 & 0.674 & 0.807 & 0.943 & 0.500 & 0.691 & 0.860 & 0.964 \\
        \midrule
        MLP & Bits        & 0.505 & 0.571 & 0.671  & 0.829 & 0.510 & 0.619 & 0.799  & 0.936 & 0.484 & 0.612 & 0.862  & 0.965 \\
        LSTM & Bits       & 0.528 & 0.636 & 0.734  & 0.856 & 0.498 & 0.648 & 0.859  & 0.947 & 0.493 & 0.633 & 0.908 & 0.963 \\
        TF & Bits         & 0.505 & 0.581 & 0.702  & 0.856 & 0.503 & 0.620 & 0.804 & 0.938 & 0.489 & 0.626 & 0.864   & 0.955 \\
        \bottomrule
    \end{tabular}
\end{table*}

\subsection{Tasks}

We consider
three tasks: concept learning, modular arithmetic, and few-shot learning with Omniglot \cite{lake2011one}. Concept learning involves in-distribution test tasks, providing an ideal scenario for meta-learning to succeed. Modular arithmetic tests both in- and out-of-distribution generalization by splitting training and testing tasks across different moduli. Few-shot classification with Omniglot introduces a more complex scenario, where limited training data leads to out-of-distribution test tasks. Here, we assess networks' sensitivity to task distribution shifts by meta-training on different Omniglot subsets.

\subsection{Data Representation}

\begin{figure}[b]
    \centering
    \includegraphics[width=\linewidth]{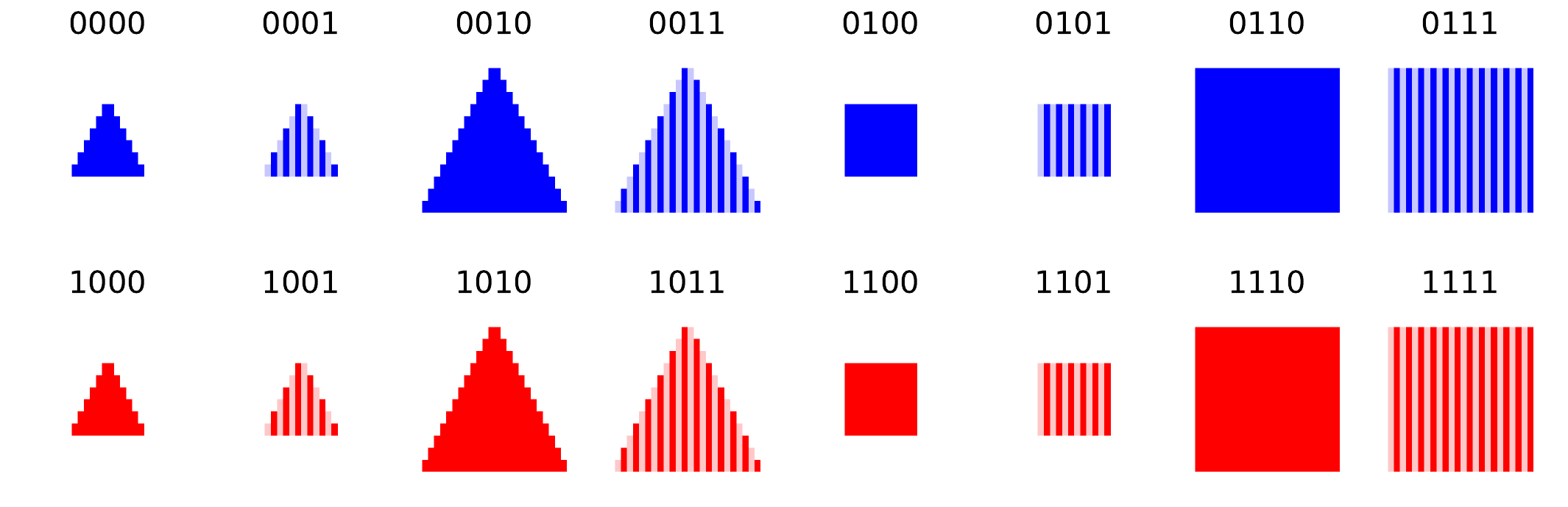}
    \caption{Input data for all 16 objects used in concept-learning with their \textit{bitstring} and \textit{image} representations.}
    \label{fig:concept_grid}
\end{figure}

\begin{figure}[b]
    \centering
    \includegraphics[width=\linewidth]{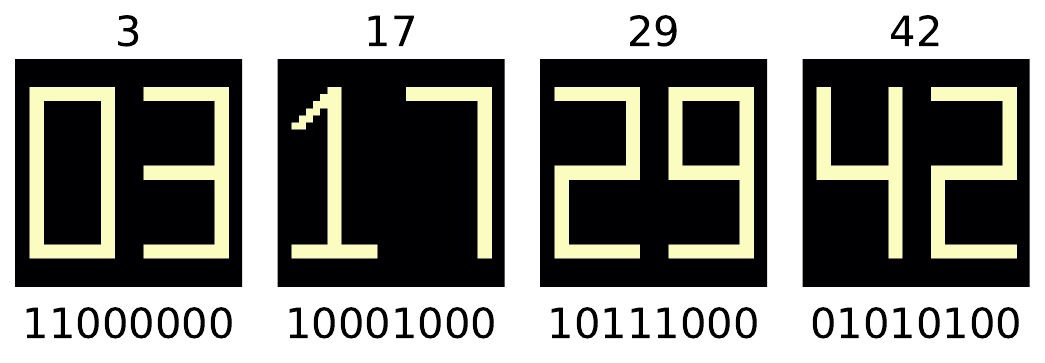}
    \caption{Input data for modular arithmetic for 4 example numbers, with \textit{number}, \textit{image}, and \textit{bitstring} representations.}
    \label{fig:number_grid}
\end{figure}

We generate two types of data representations: 32×32 images and bitstring encodings. For concept learning, each concept is represented by a 4-bit feature vector. These features include attributes such as color (red or blue), shape (square or triangle), size (big or small), and pattern (striped or solid). We visualize features as RGB images for input to the networks (see Figure~\ref{fig:concept_grid}). For modular arithmetic, input numbers are encoded as 8-bit binary strings and synthetically generated images (see Figure~\ref{fig:number_grid}). For our few-shot learning experiments with Omniglot \cite{lake2011one}, we consider exclusively the image data, downsampled to 32x32 for consistency across tasks. 

MLPs flatten the image input and process bitstrings as floating-point vectors. CNNs operate exclusively on image inputs. LSTMs and Transformers divide each 32×32 image into 4×4 patches, resulting in sequences of 64 tokens.

\subsection{Meta-Testing and Control Conditions}
We generate 100 random tasks for 10 different seeds and meta-test each of the 10 different models for each architecture class on every seed. We compare to a baseline of a random initialization using the same 10 architectures and fit 1, 10, 50, 100 and 200 steps of AdamW with a learning rate of 0.001 and a weight decay of 0.01. All tasks had converged or began to overfit after 200 steps. We use mean square error (MSE) loss as a performance metric for modular arithmetic and prediction accuracy for concept learning and Omniglot.

\section{Results}

\begin{table*}[t]
    \centering
    \caption{Average MSE for the Odd-Even Modular Arithmetic Task. Meta-Val reports training moduli while Meta-Test reports test moduli. The table compares performance under meta-learning (\textbf{M1}) and random initialization conditions with 1 and 10 steps of AdamW (\textbf{R1}, \textbf{R10}). All 95\% CIs are below 0.05 for \textbf{M1}, 0.5 for \textbf{R1}, and 0.3 for \textbf{R10}.}
    \label{tab:mod2}
    \begin{tabular}{llcc|cc|cc}
        \toprule
        & & \multicolumn{2}{c}{$n_{\text{support}} = 20$} & \multicolumn{2}{c}{$n_{\text{support}} = 40$} & \multicolumn{2}{c}{$n_{\text{support}} = 100$} \\
        \cmidrule(lr){3-4} \cmidrule(lr){5-6} \cmidrule(lr){7-8}
        \textbf{Arch.} & \textbf{Data} & \textbf{Meta-Val} & \textbf{Meta-Test} & \textbf{Meta-Val} & \textbf{Meta-Test} & \textbf{Meta-Val} & \textbf{Meta-Test} \\
        \midrule
        MLP \textbf{(M1)} & Image         & 0.310 & 0.978 & 0.246 & \textbf{0.895} & 0.211 & 0.984 \\
        CNN \textbf{(M1)} & Image       & 0.441 & 1.205 & 0.350 & 1.042 & 0.293 & \textbf{0.966} \\
        LSTM \textbf{(M1)} & Image      & \textbf{0.123} & \textbf{0.873} & \textbf{0.116} & 1.048 & \textbf{0.113} & 1.300 \\
        TF \textbf{(M1)} & Image & 0.884 & 2.476 & 0.527 & 1.731 & 0.460 & 1.414 \\
        \midrule
        MLP \textbf{(M1)} & Bits        & 0.404 & 1.021 & 0.315 & 1.041 & 0.247 & \textbf{1.050} \\
        LSTM \textbf{(M1)} & Bits      & \textbf{0.066} & \textbf{0.997} & \textbf{0.050} & \textbf{0.813} & \textbf{0.045} & 2.057 \\
        TF \textbf{(M1)} & Bits & 1.218 & 2.095 & 0.888 & 1.351 & 0.769 & 1.566 \\
        \midrule
        \midrule
        MLP \textbf{(R1)} & Image         & 40.482 & 39.134 & 35.161 & 36.544 & 32.130 & 34.546 \\
        CNN \textbf{(R1)} & Image       & 40.001 & 41.962 & 41.696 & 37.731 & 37.666 & 37.431 \\
        LSTM \textbf{(R1)} & Image      & 12.437 & 21.215 & 18.155 & 20.419 & 15.191 & 19.602 \\
        TF \textbf{(R1)} & Image & 19.195 & 16.217 & 16.014 & 17.809 & 18.164 & 16.270 \\
        \midrule
        MLP \textbf{(R1)} & Bits        & 38.627 & 35.308 & 38.724 & 39.583 & 36.076 & 37.420 \\
        LSTM \textbf{(R1)} & Bits      & 7.012 & 10.377 & 9.211 & 12.000 & 8.387 & 12.347 \\
        TF \textbf{(R1)} & Bits & 16.298 & 16.145 & 13.475 & 13.885 & 12.355 & 16.443 \\
        \midrule
        \midrule
        MLP \textbf{(R10)} & Image         & 35.497 & 35.056 & 29.221 & 28.507 & 22.461 & 24.813 \\
        CNN \textbf{(R10)} & Image       & 33.965 & 35.403 & 33.180 & 31.533 & 26.195 & 26.609 \\
        LSTM \textbf{(R10)} & Image      & 11.585 & 12.637 & 11.565 & 8.870 & 8.722 & 9.965 \\
        TF \textbf{(R10)} & Image & 11.837 & 8.843 & 8.259 & 9.156 & 8.680 & 8.753 \\
        \midrule
        MLP \textbf{(R10)} & Bits        & 33.066 & 31.307 & 28.354 & 29.302 & 24.694 & 25.449 \\
        LSTM \textbf{(R10)} & Bits      & 3.763 & 5.252 & 6.022 & 5.956 & 4.071 & 4.665 \\
        TF \textbf{(R10)} & Bits & 8.507 & 8.415 & 8.389 & 6.834 & 7.026 & 9.658 \\
        \bottomrule
    \end{tabular}
\end{table*}

\begin{table*}[t]
    \centering
    \caption{Average accuracy for the 20-way 5-shot classification task on the Omniglot experiment. The table compares performance under meta-learning (\textbf{M1}) and random initialization conditions with 1, 10, 50, 100, and 200 steps of AdamW (\textbf{R1}, \textbf{R10}, \textbf{R50}, \textbf{R100}, \textbf{R200}). Accuracy is reported for the full training set (\textbf{All}) and for subsets, including \textbf{Ancient} (12 alphabets), \textbf{Asian} (11 alphabets), \textbf{Middle Eastern} (7 alphabets), and \textbf{European} (5 alphabets). All 95\% CIs are below 0.005.}
    \label{tab:few}
    \begin{tabular}{lccccc|ccccc}
        \toprule
        & \multicolumn{1}{c}{\textbf{R1}} & \multicolumn{1}{c}{\textbf{R10}} & \multicolumn{1}{c}{\textbf{R50}} & \multicolumn{1}{c}{\textbf{R100}} & \multicolumn{1}{c}{\textbf{R200}}  & \multicolumn{5}{c}{\textbf{M1}} \\
        \cmidrule(lr){2-2} \cmidrule(lr){3-3} \cmidrule(lr){4-4} \cmidrule(lr){5-5} \cmidrule(lr){6-6} \cmidrule(lr){7-11}
        \textbf{Arch.} & \textbf{N/A} & \textbf{N/A} & \textbf{N/A} & \textbf{N/A} & \textbf{N/A}  & \textbf{All} & \textbf{Ancient} & \textbf{Asian} & \textbf{Middle Eastern} & \textbf{European} \\
        \midrule
        MLP & 0.056 & 0.149 & 0.270 & 0.279 & 0.285 & 0.753 & 0.601 & 0.605 & 0.753 & 0.753 \\
        CNN & 0.074 & 0.403 & 0.707 & 0.726 & 0.736 & 0.949 & 0.898 & 0.905 & 0.945 & 0.948 \\
        LSTM & 0.053 & 0.056 & 0.070 & 0.078 & 0.083 & 0.260      & 0.107 & 0.146 &  0.256 & 0.256 \\
        TF & 0.053 & 0.075 & 0.115 & 0.118 & 0.119 & 0.896 & 0.554 & 0.614 & 0.896 & 0.896 \\
        \bottomrule
    \end{tabular}
\end{table*}

For each of our three tasks, we evaluate the roles of architecture, data representation, and training algorithms.

\subsection{Concept Learning}
This experiment involves learning concepts based on objects that have 4 features, represented as a 4-bit vector or an image. A concept, such as $f_1(x) = 0 \land f_3(x) = 1$, assigns true or false labels to the 16 possible objects based on whether they satisfy the concept.  Following the procedure from \citeA{marinescu2024distilling}, we generate concepts from a probabilistic context-free grammar \cite{goodman2008rational}. Concepts are sampled for meta-training with variable support sizes and meta-tested on new concepts with support sizes of 5, 10, and 15. We evaluate both meta-learned models and randomly initialized models with variable gradient step counts (see Table~\ref{tab:concept}). The meta-learned models perform comparably across all architectures and both data representations, with accuracy improving as support size increases. In contrast, randomly initialized models trained with 200 steps show significantly greater performance variation across architectures and data representations, indicating that meta-learning can not only enhance task performance but also reduce the influence of architectural and representational biases on model behavior under ideal conditions.

\subsection{Modular Arithmetic}
We frame modular arithmetic as a non-linear regression task over the integer domain $[0, 100)$ in this experiment. The goal is to infer the underlying function given noisy samples from some modulus $m$. We explore two versions of the task: \emph{Odd-Even}, where we meta-train with odd moduli and meta-test with even moduli in the range $[1, 40]$, and \emph{20-20}, where we meta-train with $[1, 20]$ and meta-test with $[21, 40]$. We sample variable support sizes during meta-training and meta-test with support set sizes 20, 40, and 100, where 20, 40 are uniformly sampled and 100 includes every integer in the domain, offering a noisy version of the true moduli function over the entire domain. Noise is injected via independent samples from a Gaussian with $\sigma = 0.1$. 

We find that this task produces variations across both data representation and architecture in meta-learned models. Meta-Val reports the same moduli seen during meta-training but with a different noise seed and Meta-Test reports unseen moduli (see Table~\ref{tab:mod2}). Meta-learned LSTMs usually perform the best across data representations and support sizes.
Every model performs reasonably well, to varying degrees, for in-distribution tasks but shows a drop in performance for out-of-distribution tasks in \emph{Odd-Even}. The same is observed in \emph{20-20}, but the drop in performance with out-of-distribution tasks is catastrophic (see Table~\ref{tab:mod2020}): all meta-trained architectures have MSE's $> 50.0$. This is likely due to the harder form of generalization that is required.

Randomly initialized architectures start to overfit after just 10 steps of AdamW across all support sizes in this task, despite AdamW's explicit regularization towards simpler solutions. We report 1-step and 10-step updates (R1, R10). These randomly initialized models are significantly outperformed by their meta-trained counterparts. Notably, large variations across architectures and data representations are for the most part eliminated when the networks are meta-trained. LSTMs far outperform the other architectures, perhaps explaining their superior performance when meta-trained.

\subsection{Few-Shot Learning}
We replicate the 20-way, 5-shot Omniglot challenge \cite{lake2011one} to explore how task distribution affects performance across architectures. Handwritten alphabets are divided into four categories: Ancient (pre-500 A.D.), Asian, Middle Eastern, and European. Each neural architecture is meta-trained on these subsets. The fictional alphabets, Futurama and Magi, are excluded from the subsets but included in the base task distribution, \emph{All}, which contains all 30 training alphabets. The training baseline, \emph{N/A}, denotes the random initialization cases (R1, R10, R50, R100, R200) where we optimize for 1, 10, 50, 100, and 200 steps. All architectures are converged beyond this point.

Meta-testing on the 20 held-out alphabets reveals drops in performance for Ancient and Asian categories across all architectures (see Table~\ref{tab:few}). CNNs retain the highest percentage of their original accuracy compared to \emph{All}, while Transformers and LSTMs suffer larger drops. Transformers generalize worse than MLPs when trained on Ancient alphabets. However, generalization performance largely recovers on Middle Eastern and European alphabets. 

MLPs and Transformers can outperform randomly initialized CNNs when meta-trained. However, CNNs exhibit desirable equivariances, making them less sensitive to distribution shifts and allowing them to achieve better random initialization performance with fewer gradient steps. The performance gap between architectures narrows considerably under meta-learning. However, LSTMs still struggle with the task despite meta-learning improving their performance.

\section{Discussion}
Neural network architectures are often designed with specific problems in mind (e.g., next-word prediction, image classification), so it is natural to expect them to perform poorly on problems they were not explicitly designed for. Indeed, we found that the standard neural networks, trained without meta-learned inductive biases, perform significantly worse when the requirements of the task were misaligned with the inductive biases of the architecture. For example, tasks involving bitstrings require an explicit understanding of positional order, which is naturally encoded in sequential architectures like LSTMs and Transformers. As might be expected from these architectural properties, these architectures learned faster and performed better than MLPs, which lack this inductive bias. Similarly, in the Omniglot task, CNNs outperformed other architectures, likely due to their useful spatial inductive biases, demonstrating their superior ability to generalize when spatial structure is critical. However, even these task-suitable inductive biases were not enough to enable models to perfectly solve these tasks.

Classical analyses of learnability in cognitive science have argued for innate cognitive structures that support rapid learning from limited data, such as Chomsky’s notion of Universal Grammar (UG) as a description of an innate device that supports efficient language acquisition \cite{chomsky1980rules}, and Fodor’s theory of domain-specific encapsulated modules \cite{fodor1983modularity}. Neural networks, by contrast, do not explicitly have these structures. However, meta-learning can embed task-specific knowledge into a network’s initial weights, allowing networks to overcome limitations in architecture or data representation when exposed to a sufficiently rich task distribution. For example, \citeA{mccoy2023modeling} showed that meta-learning enables networks to learn linguistic patterns from a few examples, mimicking UG-like rapid learning. Similarly, \citeA{zintgraf2019fast} found that networks can develop task-specific specializations with minimal data, resembling Fodor’s idea of modularity. Our findings reinforce these results, showing that meta-learning vastly improves few-shot learning performance and reduces variations across architectures, suggesting that certain kinds of innate knowledge can be implicitly expressed in neural networks \cite{elman1996rethinking} and that architecture is a weak constraint on what a neural network can do.

Meta-learning still lacks the ability to distill stronger forms of generalization. Humans excel at both interpolation (learning within the range of observed examples) and extrapolation (generalizing beyond those examples). However, in modular arithmetic, we showed some generalization capabilities when fitting moduli between known moduli (interpolation) in \emph{Odd-Even} but catastrophic performance for \emph{20-20}, where the testing moduli were far outside the training task domain (extrapolation). Meta-learning can perhaps then be viewed as a ``blind'' optimization process \cite{hasson2020direct}, where we can only distill structures present in the training task distribution but not outside of it. To achieve stronger forms of generalization, we may need regularity that comes from outside of pure optimization alone, which is where architectural constraints come to play. In our Omniglot experiments, the CNN was the least sensitive to shifts in the training task distribution, demonstrating how architecture can make up the difference for what might not be available in the training data. Alternatively, we can consider techniques like reinforcement learning, that explicitly incentivize stronger forms of generalization \cite{akkaya2019solving}, or variations on meta-learning that encourage the training algorithm to find structures outside of what the data can offer \cite{irie2024neural}. 

\section{Conclusion}
While neural architectures do impose constraints on the kinds of problems neural networks can solve, these constraints are weak relative to the inductive biases afforded by initial weights. Meta-learning offers a path to distilling task-specific knowledge that is less influenced by the architecture and data representation than typical training regimes. We conclude that the flexibility of initial weights is extensive enough to override the influence of architecture in some settings, but substantial architectural differences persist when extensive generalization beyond the meta-task distribution is required. 

\section{Acknowledgments}
We would like to thank Jake Snell and Logan Nelson for their valuable feedback and discussions that contributed to this work. GB and TLG acknowledge grant DBI-2229929 from the National AI Institute for Artificial and Natural Intelligence (ARNI) and grant N00014-23-1-2510 from the Office of Naval Research, both of which supported this work. MG acknowledges support from the Princeton AI Teaching Fellowship. IM is supported by the NRT-HDR: FUTURE grant. We also gratefully acknowledge the computational resources provided by the Della high-performance computing cluster at Princeton University, which were essential for meta-training the large suite of models presented in this study.

\bibliographystyle{apacite}

\setlength{\bibleftmargin}{.125in}
\setlength{\bibindent}{-\bibleftmargin}

\bibliography{references}

\onecolumn
\section{Appendix}

\subsection{Additional Results: Modular Arithmetic}
Table~\ref{tab:mod2020} shows the average MSE errors over 10 sampled architectures for 10 random seeds for the 20-20 task, where the moduli were split into a 1-20 training group and a 21-40 testing group. This task requires more robust forms of generalization since the test task distribution is much farther away than the Odd-Even task reported in the main text. We found that performance across all meta-trained models was very poor on unseen tasks and comparable to the Odd-Even task on validation tasks. This suggests that meta-learning failed to distill the knowledge necessary to generalize to unseen moduli functions, limiting its ability to fit moduli beyond those encountered or closely related to those seen during training.

\begin{figure*}[t]
    \centering
    \includegraphics[width=\linewidth]{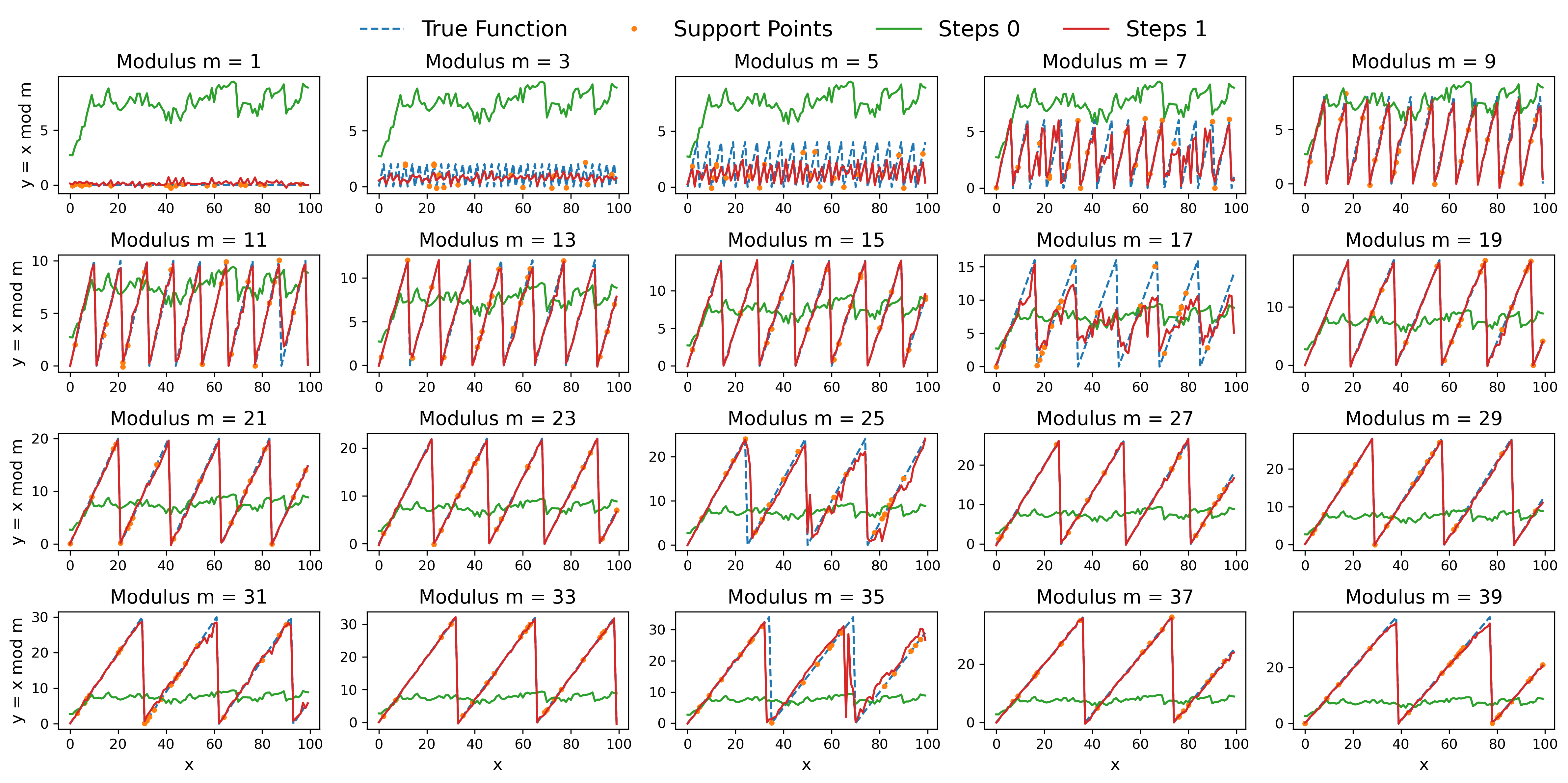}
    \caption{Visualization of Meta-Validation curve fitting for Odd-Even task using a meta-trained LSTM with image inputs and 20 support points. LSTMs were meta-trained on odd moduli (shown above) and meta-tested on even moduli. Steps 0 denotes the function before observing the support set (green). Steps 1 (red) shows the adaptation after 1 step of gradient descent. True function (blue) denotes the ground truth moduli function.}
    \label{fig:even_odd_b}
\end{figure*}

\begin{figure*}[t]
    \centering
    \includegraphics[width=\linewidth]{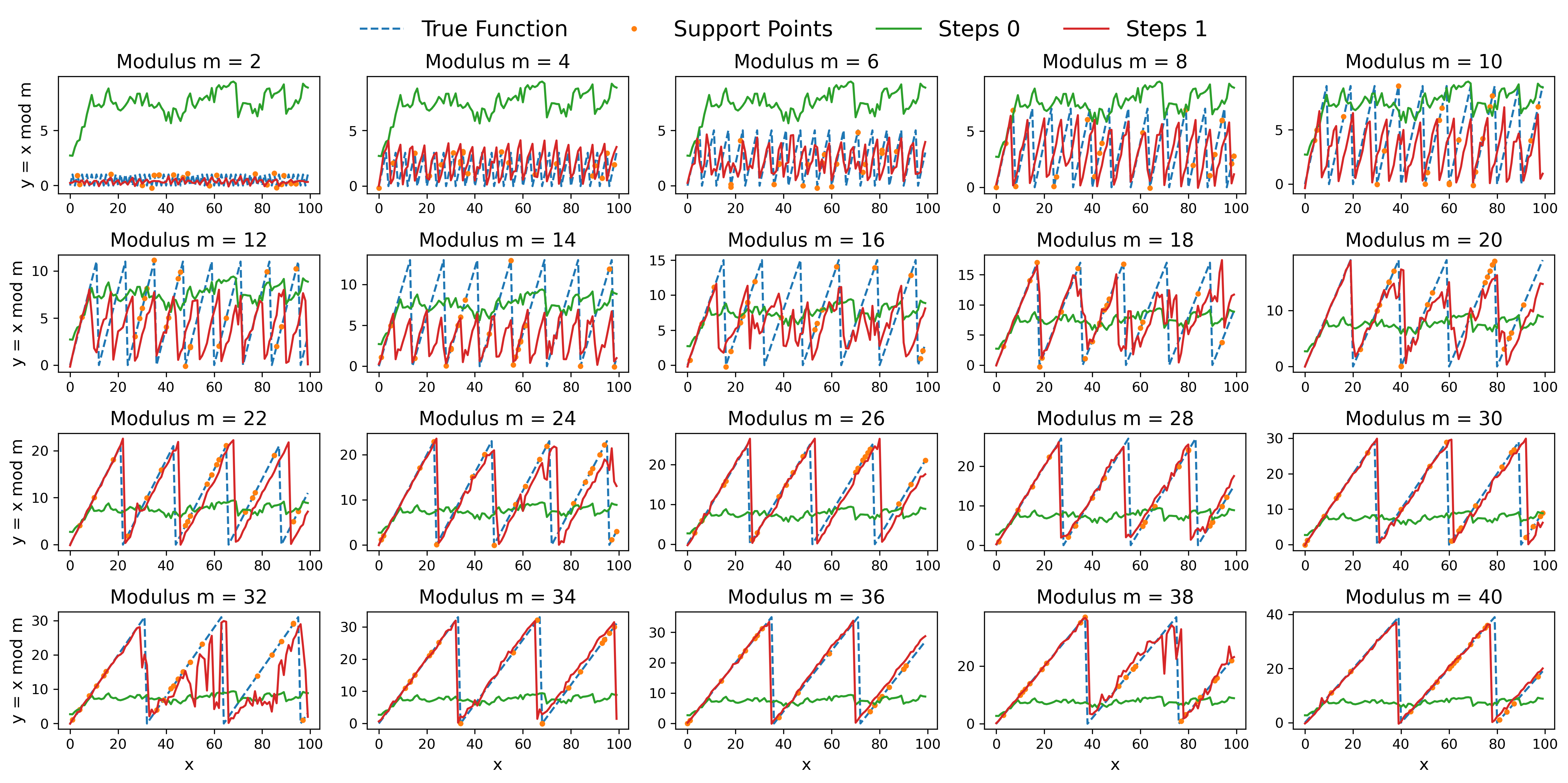}
    \caption{Visualization of Meta-Test curve fitting for Odd-Even task using a meta-trained LSTM with image inputs and 20 support points. LSTMs were meta-trained on odd moduli and meta-tested on even moduli (shown above).}
    \label{fig:even_odd_a}
\end{figure*}

\begin{figure*}[t]
    \centering
    \includegraphics[width=\linewidth]{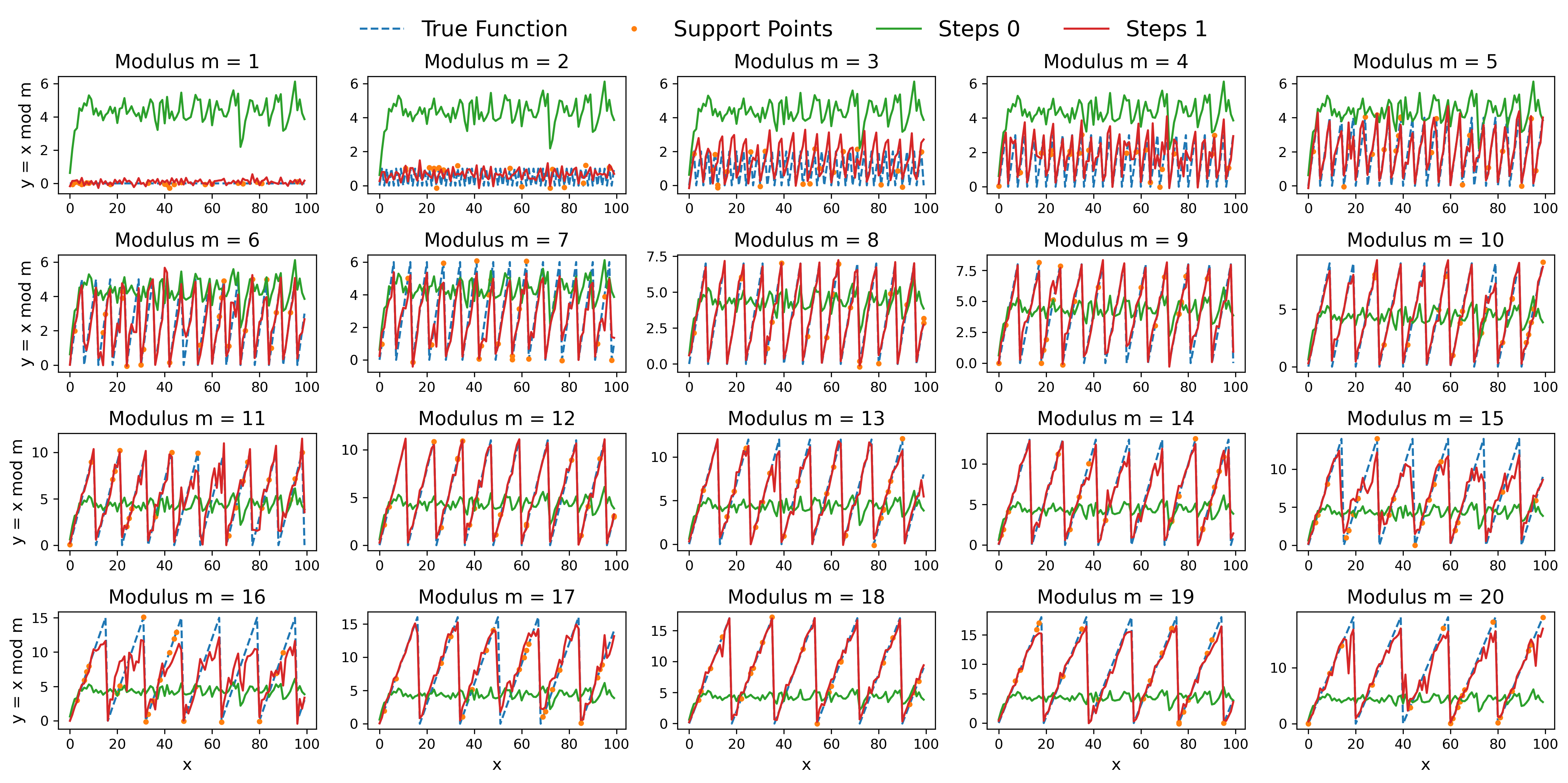}
    \caption{Visualization of Meta-Validation curve fitting for 20-20 task using a meta-trained LSTM with image inputs and 20 support points. LSTMs were meta-trained on moduli 1-20 (shown above) and meta-tested on moduli 21-40.}
    \label{fig:2020_b}
\end{figure*}

\begin{figure*}[t]
    \centering
    \includegraphics[width=\linewidth]{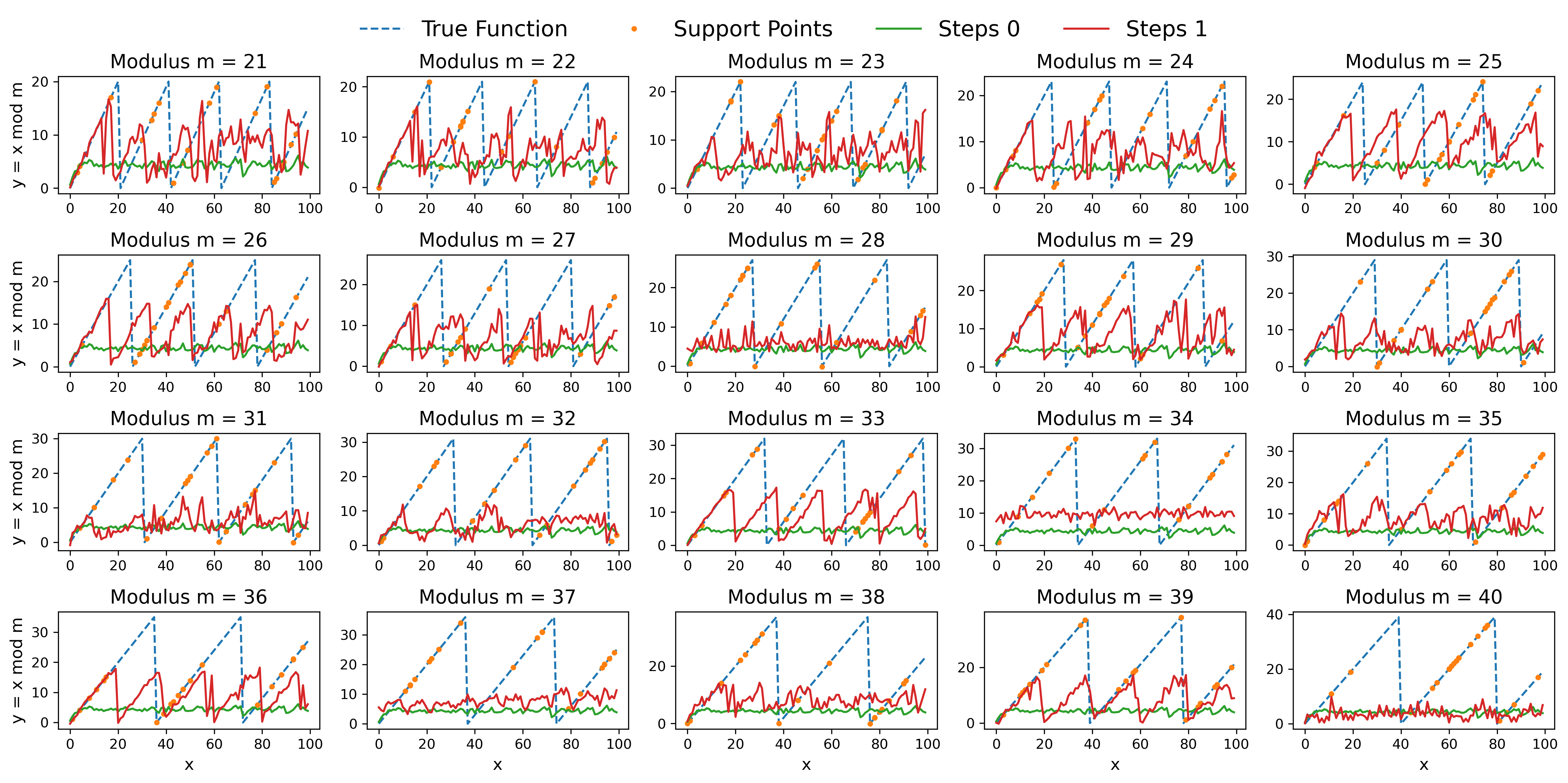}
    \caption{Visualization of Meta-Test curve fitting for 20-20 task using a meta-trained LSTM with image inputs and 20 support points. LSTMs were meta-trained on moduli 1-20 and meta-tested on moduli 21-40 (shown above).}
    \label{fig:2020_a}
\end{figure*}

In Figures~\ref{fig:2020_b} and~\ref{fig:2020_a}, we visualize representative curves for a single LSTM trained with image data and given 20 support points on the 20-20 task. Performance is near perfect on in-distribution tasks but the LSTM attempts to fit curves that resemble training moduli despite significant signal to support a different function. Figures~\ref{fig:even_odd_b} and~\ref{fig:even_odd_a} show the same curves but for the Odd-Even version of the modular arithmetic task that we report in the main text.

\begin{table*}[h]
    \centering
    \begin{tabular}{llcc|cc|cc}
        \toprule
        & & \multicolumn{2}{c}{$n_{\text{support}} = 20$} & \multicolumn{2}{c}{$n_{\text{support}} = 40$} & \multicolumn{2}{c}{$n_{\text{support}} = 100$} \\
        \cmidrule(lr){3-4} \cmidrule(lr){5-6} \cmidrule(lr){7-8}
        \textbf{Arch.} & \textbf{Data} & \textbf{Meta-Val} & \textbf{Meta-Test} & \textbf{Meta-Val} & \textbf{Meta-Test} & \textbf{Meta-Val} & \textbf{Meta-Test} \\
        \midrule
        MLP \textbf{(M1)} & Image         & 0.224 & 62.303 & 0.137 & 52.401 & 0.085 & 53.860 \\
        CNN \textbf{(M1)} & Image       & 0.291 & 65.640 & 0.204 & 59.214 & 0.127 & 55.021 \\
        LSTM \textbf{(M1)} & Image      & \textbf{0.072} & 89.356 & \textbf{0.049} & 87.886 & \textbf{0.042} & 92.151 \\
        TF \textbf{(M1)} & Image        & 0.265 & 79.551 & 0.203 & 77.452 & 0.163 & 84.379 \\
        \midrule
        MLP \textbf{(M1)} & Bits        & 0.248 & 60.441 & 0.165 & 52.989 & 0.099 & 52.352 \\
        LSTM \textbf{(M1)} & Bits      & \textbf{0.027} & 99.372 & \textbf{0.021} & 103.318 & \textbf{0.018} & 112.794 \\
        TF \textbf{(M1)} & Bits         & 0.275 & 71.229 & 0.270 & 75.373 & 0.200 & 77.683 \\

        \midrule
        \midrule
        MLP \textbf{(R1)} & Image         & 27.606 & 66.857 & 23.333 & 55.392 & 22.971 & 45.168 \\
        CNN \textbf{(R1)} & Image       & 28.265 & 63.105 & 24.337 & 51.644  & 22.828 & 43.167 \\
        LSTM \textbf{(R1)} & Image      & 5.229 & 56.915 & 6.812 & 51.979 & 7.534 & 50.939 \\
        TF \textbf{(R1)} & Image        & 11.432 & 54.120 & 11.611 & 50.490 & 13.394 & 47.850 \\
        \midrule
        MLP \textbf{(R1)} & Bits        & 26.195 & 61.145 & 22.190 & 50.674 & 21.967 & 40.578  \\
        LSTM \textbf{(R1)} & Bits      & 3.895 & 47.907 & 4.837 & 40.954 & 4.258 & 37.623 \\
        TF \textbf{(R1)} & Bits         & 8.872 & 47.622  & 9.543 & 43.548 & 9.126 & 42.618 \\
        \midrule
        \midrule
        MLP \textbf{(R10)} & Image         & 24.947 & 60.052 & 19.750 & 47.102 & 16.940 & 31.091 \\
        CNN \textbf{(R10)} & Image       & 24.992 & 55.891 & 20.042 & 43.738 & 16.926 & 29.164 \\
        LSTM \textbf{(R10)} & Image      & 4.421 & 57.613 & 4.386 & 50.533 & 4.892 & 49.009 \\
        TF \textbf{(R10)} & Image       & 6.381 & 49.464 & 5.687  & 45.311 & 5.393 & 39.720 \\
        \midrule
        MLP \textbf{(R10)} & Bits        & 22.877 & 52.845 & 17.809 & 39.234 & 15.418 & 25.106 \\
        LSTM \textbf{(R10)} & Bits      & 2.331 & 45.800 & 2.525 & 35.331 & 2.174 & 21.841 \\
        TF \textbf{(R10)} & Bits        & 6.782 & 47.109 & 5.671 & 42.674 & 4.163 & 38.185  \\
        \bottomrule
    \end{tabular}
        \caption{Average MSE for the 20-20 Modular Arithmetic Task. Meta-Val reports training moduli while Meta-Test reports test moduli. The table compares performance under meta-learning (\textbf{M1}) and random initialization conditions with 1 and 10 steps of AdamW (\textbf{R1}, \textbf{R10}). All 95\% confidence intervals (CIs) are below 0.05 on Meta-Val and 0.5 on Meta-Test for \textbf{M1}, below 0.5 for \textbf{R1}, and below 0.3 for \textbf{R10}.}
        \label{tab:mod2020}
\end{table*}

\end{document}